%% file: spark.tex
\documentclass[letterpaper, 10 pt, conference]{ieeeconf}  
\IEEEoverridecommandlockouts                              
\input{misc/packages}

\usepackage[noadjust]{cite}
\usepackage{amsmath,amssymb,amsfonts}
\usepackage{algorithmic}
\usepackage{graphicx}

\usepackage{subcaption}
\usepackage{textcomp}
\usepackage{xcolor}

\usepackage{hyperref}
\let\labelindent\relax

\usepackage{enumitem}
\usepackage{multirow}
\usepackage{balance}
\usepackage{hhline}

\usepackage{tabularx}

\title{\LARGE \bf
SPARK-Remote: A Cost-Effective System for Remote Bimanual \\ Robot Teleoperation
}

\author{Adam Imdieke and Karthik Desingh
}

\let\oldtwocolumn\twocolumn
\renewcommand\twocolumn[1][]{%
    \oldtwocolumn[{#1}{
        \centering
        \includegraphics[width=1.0\textwidth]{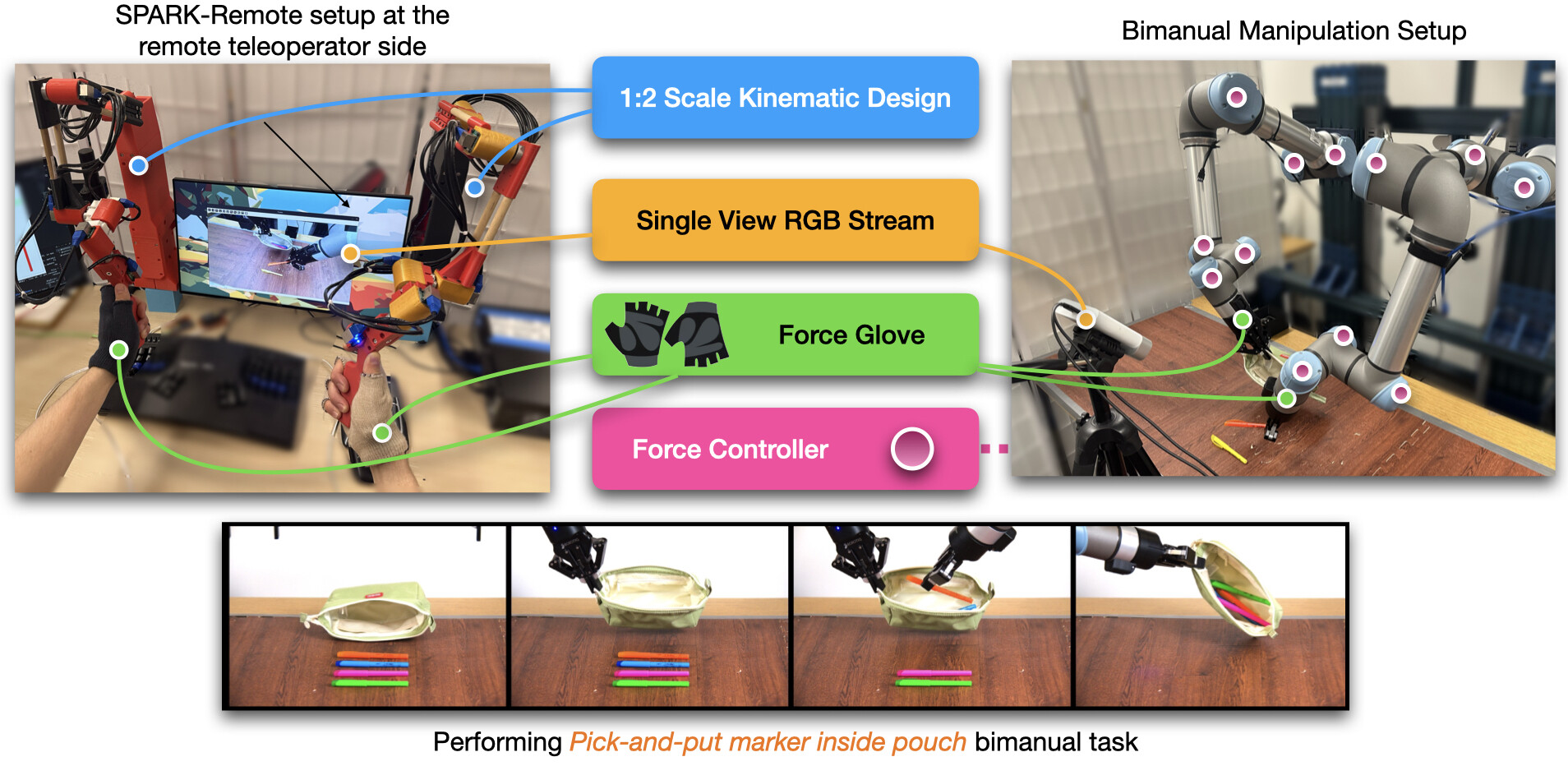}
        \captionof{figure}{The SPARK-Remote system (left) is a low-cost bimanual teleoperation system with kinematics scaled to match the bimanual robot manipulator (right). This work extends the SPARK setup by incorporating a single-view RGB camera stream, a force glove for haptic feedback, and a force controller to prevent high contact forces. We evaluate SPARK's effectiveness against other teleoperation technologies and its own variants across various bimanual manipulation tasks. Our study highlights the challenges of remote bimanual teleoperation and demonstrates how haptic feedback and force limting can addresses them. The cost per scaled arm, including  one force glove, is approximately \$200 USD.}
        \label{fig:intro}
    }]
}

\usepackage{etoolbox}
\makeatletter
\patchcmd{\@makecaption}
  {\scshape}
  {}
  {}
  {}
\makeatother

\begin{document}
\maketitle



\input{sections/0_Abstract/Abstract}

\input{sections/1_Introduction/Introduction}
\input{sections/1_Introduction/RelatedWorks}

\input{sections/2_Methods/Hardware_Design}

\input{sections/2_Methods/Setup}
\input{sections/2_Methods/Tasks}
\input{sections/3_Results/Discussion}
\input{sections/4_Conclusion/Conclusion}

\input{sections/4_Conclusion/Acknowledgments}

\bibliographystyle{IEEEtran}
\bibliography{spark}

\end{document}

%% file: misc/packages.tex
\usepackage{graphicx}
\usepackage{float}

%% file: sections/0_Abstract/Abstract.tex
\begin{abstract}
Robot teleoperation enables human control over robotic systems in environments where full autonomy is challenging. Recent advancements in low-cost teleoperation devices and VR/AR technologies have expanded accessibility, particularly for bimanual robot manipulators. However, transitioning from in-person to remote teleoperation presents challenges in task performance. 
We introduce SPARK, a kinematically scaled, low-cost teleoperation system for operating bimanual robots. Its effectiveness is compared to existing technologies like the 3D SpaceMouse and VR/AR controllers. We further extend SPARK to SPARK-Remote, integrating sensor-based force feedback using haptic gloves and a force controller for remote teleoperation. We evaluate SPARK and SPARK-Remote variants on 5 bimanual manipulation tasks which feature operational properties - positional precision, rotational precision, large movements in the workspace, and bimanual collaboration -  to test the effective teleoperation modes. Our findings offer insights into improving low-cost teleoperation interfaces for real-world applications.
For supplementary materials, additional experiments, and qualitative results, visit the project webpage: \url{https://bit.ly/41EfcJa}.

\end{abstract}

%% file: sections/1_Introduction/Introduction.tex
\section{Introduction}
Robot teleoperation is a fundamental approach for leveraging mechanical systems in scenarios that are challenging to automate and where direct human operation is infeasible. Recently, there has been a surge in low-cost teleoperation devices and the broader accessibility of emerging technologies, such as VR/AR, particularly for bimanual robot manipulators. This growth is driven by the intuitive nature of bimanual manipulation, making teleoperation a key method for imparting autonomy to robots in these tasks. As remote teleoperation emerges as a viable option for operators worldwide, there is a pressing need to understand the transition from in-person to remote teleoperation, particularly for bimanual robots, and to examine the factors influencing the effectiveness of these systems in performing complex manipulation tasks. In this paper, we introduce \textbf{SPARK} (\textbf{S}caled \textbf{P}roportional \textbf{A}rms for \textbf{R}obot \textbf{K}inematics), a low-cost teleoperation device, and evaluate its effectiveness in controlling a bimanual robot compared to existing technologies. Additionally, we develop SPARK-Remote to adapt SPARK for remote teleoperation and evaluate its effectiveness in controlling a bimanual robot manipulator.

Various user interfaces have been employed to teleoperate robot manipulators, primarily for collecting expert demonstrations, with differing levels of performance and intuitiveness. Among these, the 3D SpaceMouse by 3Dconnexion is a widely used option \cite{diff, shared, chu2023bootstrapping, gello}, alongside VR/AR controllers \cite{peract, chu2023bootstrapping, seo2023deep}. More recently, scaled kinematic platforms have gained traction \cite{aloha, mobile-aloha, gello, kim2023training, yangace}, with our SPARK system belonging to this category. In this work, we build upon the study by GELLO~\cite{gello} to evaluate how a scaled kinematic platform compares to other teleoperation technologies—such as the 3D SpaceMouse and VR/AR controllers—in executing bimanual manipulation tasks that demand varying operational degrees of \textit{positional precision}, \textit{rotational precision}, \textit{large movements}, and \textit{bimanual collaborations}, as well as their combinations.

Finally, we extend the SPARK setup from in-person to remote teleoperation, incorporating force feedback with a force glove on the operator’s end and force controller on the robot’s end. We evaluate their impact on task performance and find that these enhancements significantly improve remote teleoperation for tasks requiring high precision, close-proximity coordination between arms, and contact-rich interactions with objects and the environment.

In this paper, we introduce SPARK, a teleoperation system leveraging scaled kinematic models and low-cost hardware (\$175/arm), incorporating a novel force glove (\$25/glove) and force controller, and extending it to remote teleoperation (SPARK-Remote).
The primary contributions of this paper are as follows:
\begin{enumerate}
    \item An open-source bimanual robot teleoperation platform, proposing a low-cost scaled kinematic device for high-precision teleoperation with a force glove and force controller for enhancing remote operation.
    \item A comprehensive evaluation of SPARK, SPARK-Remote, and its variants across five carefully designed bimanual tasks, emphasizing \textit{positional and rotational precision, large movements, and bimanual collaboration}.
    \item Extensive analysis of collected data to assess the effectiveness of each SPARK-Remote variant in relation to key operational properties of the bimanual tasks.
\end{enumerate} 

%% file: sections/1_Introduction/RelatedWorks.tex
\section{Related Work}
\label{sec:realted_work_chapter}

\subsection{Teleoperation Methods for Robot Manipulators}
Teleoperation devices are typically categorized as unilateral or bilateral~\cite{tele}. Unilateral systems use a leader-follower model, where data flows one way from the leader to the follower without feedback. Bilateral systems enable two-way data flow, providing feedback to the operator. While bilateral teleoperation is ideal, it is often prohibitively expensive, with devices like the Omega7 costing around \$40,000. A mirrored 1-to-1 setup with a high-capacity manipulator like the UR5e would cost approximately \$30,000 per arm. Due to these high costs, unilateral teleoperation remains a more practical and affordable alternative.  

Unilateral teleoperation takes various forms, including generic 6DoF controllers like VR controllers~\cite{VR, peract} and the SpaceMouse~\cite{diff}. These interfaces use Cartesian movements with inverse kinematics and often lack force feedback, limiting the operator’s awareness of end-effector forces.

\subsection{Haptic Feedback in Teleoperation}
Haptic feedback during teleoperation enables the operator to sense the force exerted by the robot. While visual observation of the end effectors allows some force estimation, it is often insufficient when dealing with rigid objects, or with low acuity images, as there is insufficient visible deformation to infer contact force from. To address this, Kamijo et al.~\cite{comp} propose a compliant teleoperation system that integrates haptic feedback into VR controllers. Their system uses a position-based compliance controller, where the VR controllers adjust vibration magnitude to reflect the applied force. Our SPARK-Remote builds on this idea to indicate the force magnitude along each axis. While Rueckert et al.~\cite{glove} use vibration motors to indicate force at the fingertip of a robotic hand, we adapt this approach to SPARK-Remote to convey force at the end effector of a robotic manipulator.

\subsection{Remote Teleoperation Methods for Robot Manipulators}
Remote teleoperation introduces additional challenges, particularly latency in sensor observations and control transmission over the network. While our system focuses on low-cost teleoperation devices, several works aim to streamline remote teleoperation using 1:1 scaled systems with haptic feedback or fully immersive VR. Zhou et al.\cite{Zhou2021ABD} propose a bilateral teleoperation control for bimanual robots, addressing time delays in remote operation. Lenz et al.\cite{9568842} develop a telemanipulation system incorporating force feedback via SenseGlove and predictive limit avoidance. Both systems are 1:1 scale replicas of the remote robotic system. Han et al.~\cite{xu2025immersive} introduce an immersive teleoperation method with bilateral haptic feedback, visual rendering, and isomorphic dexterous arm and hand control. In contrast, our approach prioritizes affordability, avoiding costly haptic devices and VR-based visual rendering while maintaining effective remote operation.


%% file: sections/2_Methods/Hardware_Design.tex
\section{SPARK Hardware Design}
\label{sec:hardware}
Here, we describe our low-cost hardware design: SPARK, the force glove add-on (used in SPARK-FG), and the force controller for the UR5e (used in SPARK-FC). 

\subsection{Scaled Kinematic Model} 
\label{sec:scaled_kinematic_model}
\begin{figure}[!t]
     \centering
     \begin{subfigure}[b]{0.46\columnwidth}
         \centering
         \includegraphics[width=1.0\columnwidth]{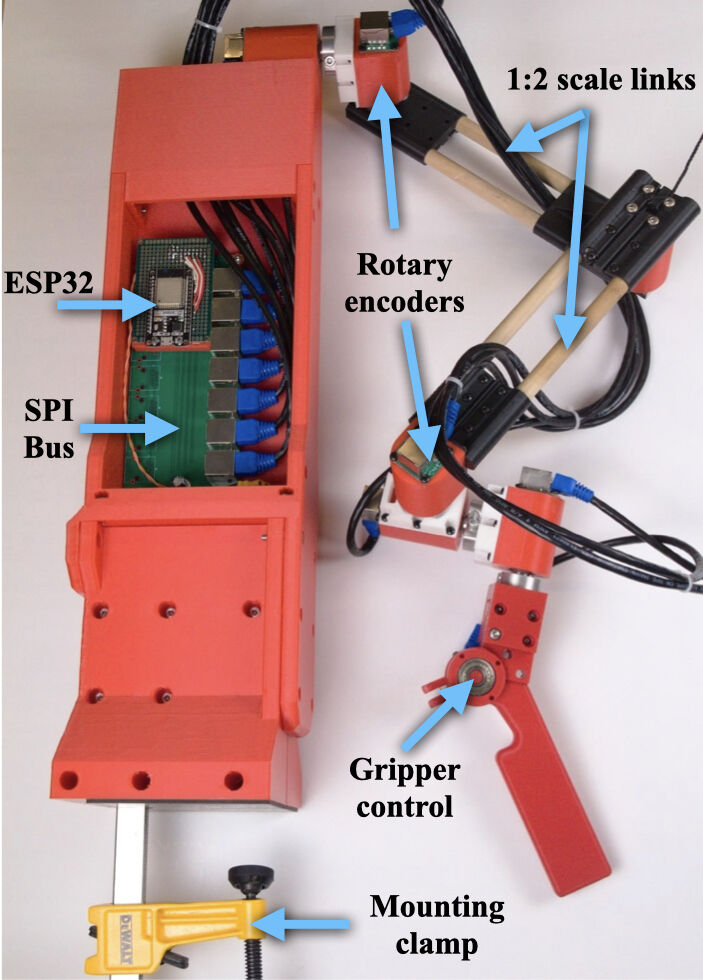}
         \caption{Single SPARK arm}
         \label{fig:hardware}
     \end{subfigure}
     \hfill
     \begin{subfigure}[b]{0.5\columnwidth}
         \centering
         \includegraphics[width=1.0\columnwidth]{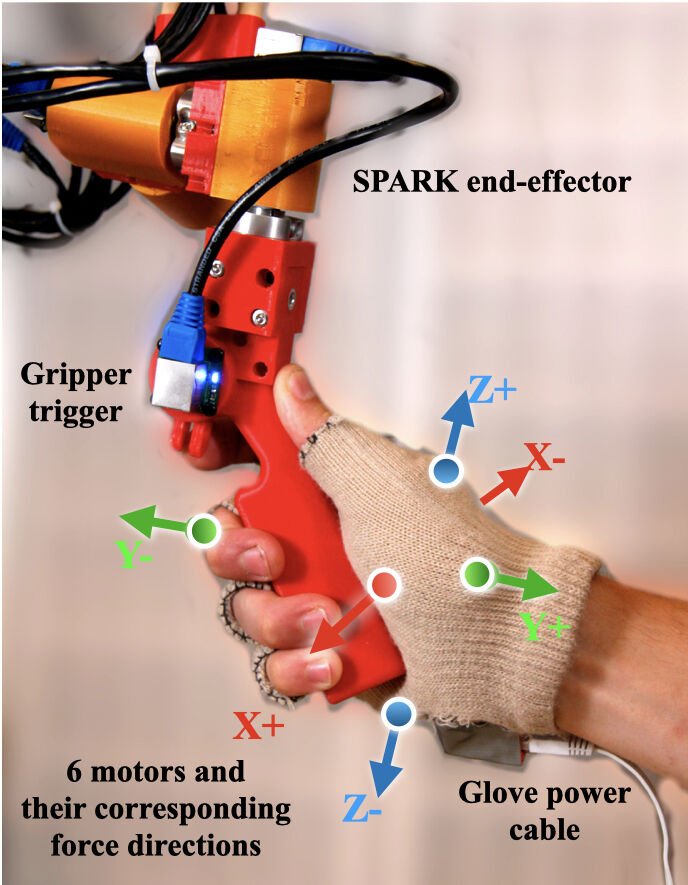}
         \caption{Force glove hardware}
         \label{fig:glove_hardware}
     \end{subfigure}
        \caption{(a) A single SPARK arm with 1:2 scale links, corresponding to the UR5e robotic arm. (b) The force glove worn by an operator interacting with the SPARK end-effector, featuring six motors that relay haptic force directions from the UR5e force/torque sensor.}
        \label{fig:hardware_spark_glove}
\end{figure}

SPARK is built from the ground up to enable affordable and intuitive bimanual teleoperation. Our design follows the trend of scaled kinematic teleoperation platforms~\cite{aloha, gello, mobile-aloha}. 
The SPARK setup mirrors our dual-arm manipulation system, featuring two Universal Robots UR5e arms mounted on a Vention frame with Robotiq 2F-85 grippers (Figure~\ref{fig:intro}). The operator moves the scaled model (leader), while the manipulator (follower) mirrors these movements in joint space. SPARK’s scaled kinematic model uses a single design for the rotational joint assembly across all joints (Figure~\ref{fig:hardware}), using ball bearings and a steel axle for high precision and low resistance. A  magnet on the end of the axel enables a 14-bit encoder to accurately measure the angle at a high sampling rate. An ESP32 microcontroller reads the angles through an SPI bus for all six joints and the gripper.
The rotary joints have minimal resistance, reducing user fatigue during long sessions. Additionally, each SPARK unit features an enable/disable pedal. SPARK’s kinematic structure mirrors the target embodiment by scaling the Denavit-Hartenberg (DH) parameters. Though designed around the UR5e, SPARK can adapt to other DH configurations by modifying joint connections. Our designs are open source, and can be found on our website. 


\subsection{Precision in Teleoperation}
\label{sec:precise_teleoperation}
A key aspect of teleoperation is the balance between coarse movements across the workspace and fine manipulation. SPARK leverages the operator’s natural arm movements to achieve this balance. SPARK’s 1:2 scale design mimics the UR5e kinematic structure, allowing the operator to move the end effectors as naturally as their own hands. Large movements cover more distance thanks to this scale, while fine adjustments can be made by slowing down and using small movements. High precision is achieved through our 14-bit magnetic encoders on each joint, offering a theoretical resolution of 0.022\textdegree (3.8e-4 rad).

\subsection{Key differences between SPARK and GELLO}
\label{sec:spark_gello}
Our work extends the findings of GELLO~\cite{gello}, where both works propose scaled kinematic controllers for the UR5. We replicate their comparison to the VR and SpaceMouse systems and validate their findings with our tasks. We go beyond the work of GELLO with our evaluation of bimanual teleoperation over a network, along with our proposed force glove and force controller. Where GELLO evaluates new users, we utilize an experienced teleoperator, so our results better represent how professional teleoperators would use these systems. While both systems use the same concept of scaled kinematic models, SPARK uses custom encoders and rotational assemblies that have very low friction and high resolution when compared to backdriving servo motors with GELLO.

\subsection{Tactile Feedback using Force Glove} 
\label{sec:tactile_feedback}
To provide feedback to the operator, we developed a haptic glove that indicates the direction and magnitude of forces on the end effector.
The glove features six coin vibration motors that translate the $(x,y,z)$ force data of the end effector’s force torque sensor to haptics for each $+/-$ axis direction. Each motor corresponds to one of the cardinal directions—up/down, left/right, front/back. Positive and negative forces activate the opposite motors, where the vibration intensity scales with force magnitude. This feedback method works particularly well with the SPARK setup, as the operator’s hands remain aligned with the end effectors, ensuring alignment of the axes of the UR's force sensors and the hand's feedback axes. (Figure \ref{fig:hardware}). The force glove allows SPARK's operators to "close the loop", using the haptic feedback to inform their next actions. This reduces the guesswork in contact rich scenarios, allowing operators to be more confident with their movements.

\subsection{Force Controller}
\label{sec:spark_force}


\input{sections/2_Methods/ForceController}


%% file: sections/2_Methods/ForceController.tex
The force controller employs an optimization-based approach to achieve joint mirroring while preventing excessive joint torques through two operating modes: SPARK mode and torque mode. Our implementation uses PyTorch, though any optimization framework can be used. The robot has $N$ joints, with coefficients: target distance $a$, step size $s$, and maximum safe torque $\tau_{\max}$ as tuning parameters. The variable $\theta$ represents the arm's current joint angles:
\begin{equation}
 \theta = \texttt{Arm.GetJointAngles()}
\end{equation}
The torque mode uses the torque $\tau_i$ from each joint $i$ is to minimize the force on the end effector. Torque mode creates a target joint angle $\hat\theta_i$, in the opposite direction of the joint torque, with the goal of reducing the torque on each joint. The overall objective for torque mode is the minimization of the MSE between $\hat{\theta}$ and $\theta$, providing the Torque loss $\mathcal{T}_{Loss}$.

\begin{align}
\hat{\theta}_i = \theta_i - (\tau_i * a) \label{eq:first} \\
\mathcal{T}_{Loss} = \sum_{i=1}^{N} (\theta_i - \hat{\theta}_i)^2 \label{eq:second}
\end{align}



$\hat{\theta}$ represents a target joint angle, allowing us to backpropagate through $({\theta}-\hat{\theta})$  and the loss will not work without it. $\hat{\theta}$ is required as the loss must create a gradient that minimizes $\theta$, and $\nabla\tau$ does not create a backpropagation path to $\theta$.

Similarly, SPARK mode attempts to reduce the difference between SPARK's joint angles and the UR5e's, giving SPARK loss as:
\begin{equation}
\mathcal{S}_{Loss} = \sum_{i=1}^{N} (\theta_i - \theta^\mathcal{S}_i)^2
\end{equation}
With small observed torques, the arm acts only in SPARK mode, where $\mathcal{S}_{Loss}$ is the dominant loss function, so the arm follows SPARK.
As the maximum observed joint torque $\max(\tau)$ increases, the controller gives less authority to mirroring, and more authority to the torque mode.
Once there is a large amount of torque, the loss is dominated by torque mode, and will only move in a direction that decreases the force. Mode ratio $\mathcal{M}$ interpolates between the torque mode and SPARK mode loss functions to enable a smooth transition. 
\begin{align}
\mathcal{M} &= \frac{\max(\tau)}{\tau_{max}} \label{eq:third} \\
\text{Loss} &= (\mathcal{T}_{Loss} \times \mathcal{M}) + (\mathcal{S}_{Loss} \times (1 - \mathcal{M})) \label{eq:fourth}
\end{align}



With the loss function now defined, we can take the gradient of our loss with respect to $\theta$ to find $\nabla\theta$, that we will treat as a velocity. The intuition behind using the gradient as a velocity vector is that moving the robot's joints in a direction that minimizes the loss will act as an update step, helping achieve the behavior defined in the loss functions.
\begin{equation}
\texttt{Arm.SetJointSpeeds}(\nabla\theta * s)
\end{equation}
From the operator’s perspective, the robot halts arm tracking upon reaching a torque threshold and only moves in directions that minimize torque after contact. More details on our implementation are available in the supplementary material, where we also apply this optimization method to condition the null space of underconstrained inverse kinematics.

%% file: sections/2_Methods/Setup.tex
\section{Experimental Setup for Teleoperation}
\label{sec:setup}

In this study, we evaluate 4 different teleoperation modes (SpaceMouse, VR controllers, SPARK, and SPARK-Remote) on our bimanual robot manipulation setup across 5 tasks showcasing the key aspects of the tasks. The key aspects show the important teleoperation aspects in each task: \textit{positional precision}, \textit{rotational precision}, \textit{motion speed}, and \textit{contact richness}. Each teleoperation method is tested by an expert operator across 10 trials for each bimanual manipulation task, providing insight into the capability of each method across each of the tasks. Moreover, the SPARK-Remote has multiple combinations designed to address the challenges in the remote teleoperation of bimanual tasks.

Our bimanual robot manipulation setup features two Universal Robots UR5e manipulators (see Figure~\ref{fig:hardware}), with a force and torque sensor at the end-effector. Our bimanual robot configuration mimics humanoids, and is not explicitly optimized for teleoperation applications. This poses several considerations: a) potential collisions between the arms - shoulder joints may collide, b) gimbal lock zones significantly reduce manipulability in a large part of the workspace, c) the grippers’ non-parallel jaws make it difficult for operators to predict grasp positioning, which can cause unintended collisions with the tabletop. These properties can result in unexpected emergency stops (e-stops, as e-stops automatically halt the arm if it violates safety parameters such as force, speed, or configuration). In our study, when an e-stop occurs, the task timer stops, and the arm may be reset to a home position. The operator can then clear the error and resume or return to the home position.

We use two SPARK units to replicate the dual-arm arrangement. An additional encoder attached to a trigger mechanism on SPARK's end effector controls the grippers. A gravity compensation system with tool balancers mounted to the ceiling is used to keep the "elbows" of the arms raised, minimizing collisions and maintaining an ergonomic configuration for the operator.

\subsection{Remote Teleoperation Network Setup}
To teleoperate remotely, we utilize ROS2 and WebRTC for our communication protocols. 
We use a wired Gigabit Ethernet connection, on an isolated network, to prevent interference. We do not impose artificial bandwidth or latency limitations unless stated. 
Our SPARK setup transmits only joint angles and receives force torque values at 30 Hz, so ROS2's transmission overhead is not a concern. 
For transmitting images, we did not use ROS2 as we found that passing 1920×1080 images resulted in high latency (500ms) and low bandwidth (10hz) for a single image channel.
We instead use a lightweight WebRTC implementation for streaming our video at much higher effective bitrates (30 Hz, 100 ms latency).
In the remote versions, the teleoperator is given a front view camera, from a similar perspective that the operator would have in-person.

\subsection{SPARK-Remote Teleoperation Methods}
In addition to using two SPARK units remotely (vision and control with a computer network), additional features to support remote teleoperation includes a) monitor displaying the camera feed, b) force gloves for the operator, and c) force compensation algorithm on the bimanual manipulator's end. To understand the effects of each of these additional features designed for the remote teleoperation, we will study the following variants of SPARK-Remote. 

\subsubsection{SPARK-Remote-Basic} This mode is a simple extension of the SPARK setup with a wired Gigabit Ethernet connection to the bimanual robotic system along with a monitor to display the robot's camera views. We tested single view vs. multiple views, including wrist camera views, and found that a single view is the most suitable for our evaluation. We found the addition of other views will take network and compute resources from the main video feed, and that switching perspectives between wrist views and a static view is disorienting.  

\subsubsection{SPARK-Remote-FG} This mode integrates the haptic/force glove to the SPARK-Remote setup, allowing operators to sense the forces on each end effector. This setup builds on the SPARK configuration, adding haptic feedback through six motors that activate based on the force magnitude on each axis $(x,y,z)$ from the bimanual robot's end effector’s force-torque sensor.

\subsubsection{SPARK-Remote-FC} 
This mode adds a force controller to manipulators in the SPARK-Remote setup as a custom motion controller for the bimanual robot. The goal of this compensation is to prevent e-stops by limiting the amount of force that can be applied by the robot.

\subsubsection{SPARK-Remote-FGC} This mode has both the force glove and the force controller (SPARK-Remote-FG + SPARK-Remote-FC).


%% file: sections/2_Methods/Tasks.tex
\section{Bimanual Tasks and Results}
\label{sec:tasks}
    
\begin{figure}[!t]
     \centering
     \begin{subfigure}[b]{1.0\columnwidth}
         \centering
         \includegraphics[width=\textwidth]{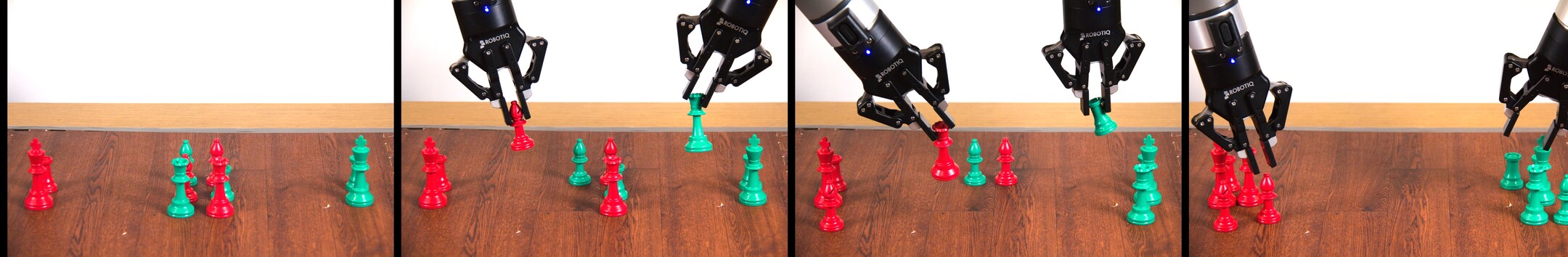}
         \caption{Pick-and-place chess pieces}
         \label{fig:chess_task}
     \end{subfigure}
     \hfill
     \begin{subfigure}[b]{1.0\columnwidth}
         \centering
         \includegraphics[width=\textwidth]{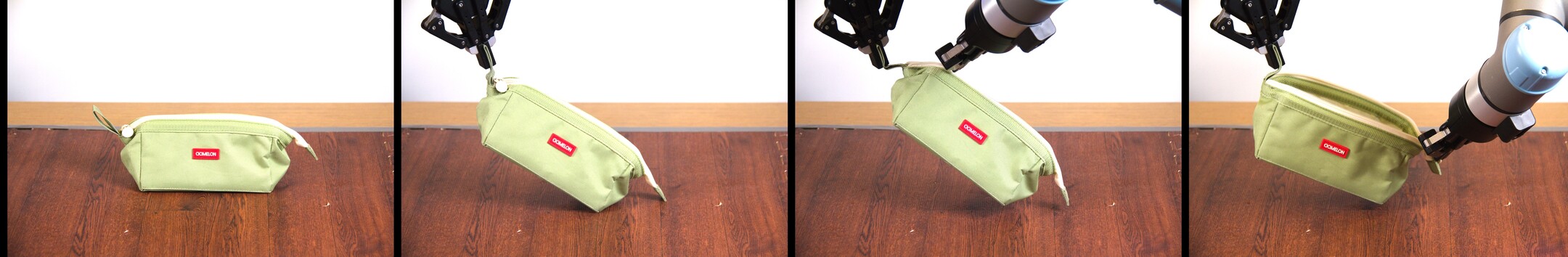}
         \caption{Unzip-and-open bag }
         \label{fig:unzip}
     \end{subfigure}
     \hfill
     \begin{subfigure}[b]{1.0\columnwidth}
         \centering
         \includegraphics[width=\textwidth]{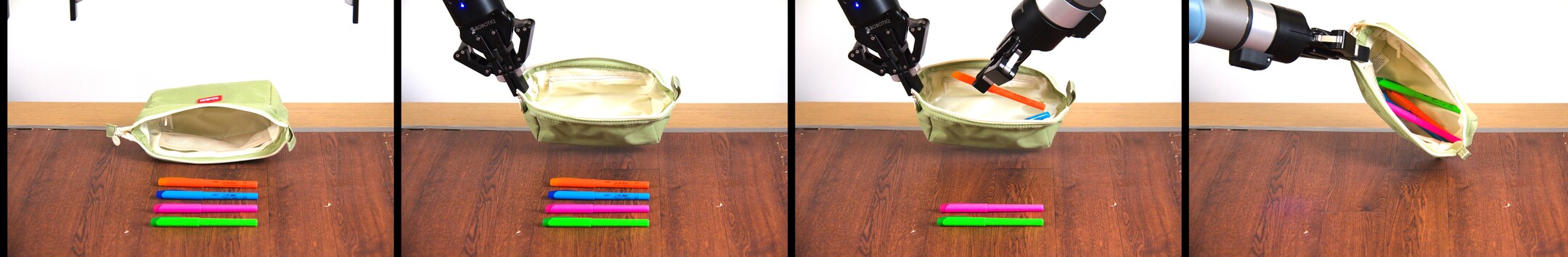}
         \caption{Pick-and-put markers in bag}
         \label{fig:markers}
     \end{subfigure}
     \hfill
     \begin{subfigure}[b]{1.0\columnwidth}
         \centering
         \includegraphics[width=\textwidth]{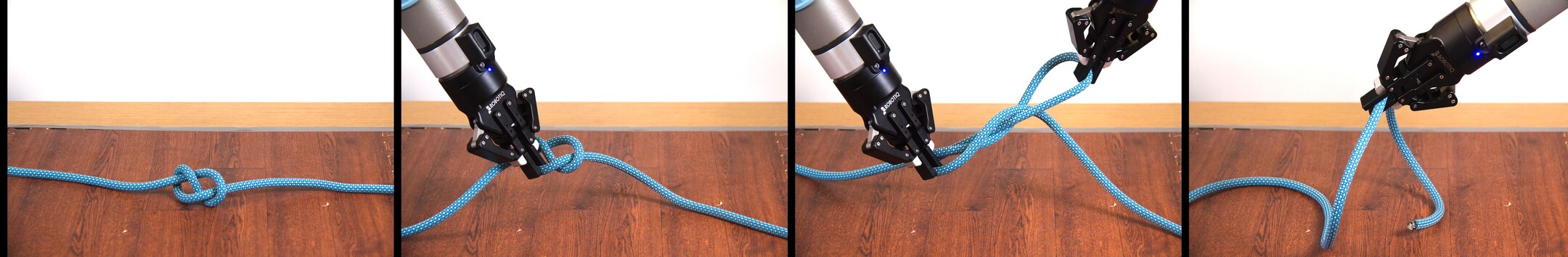}
         \caption{Untie a knot}
         \label{fig:untie}
     \end{subfigure}
    \hfill
     \begin{subfigure}[b]{1.0\columnwidth}
         \centering
         \includegraphics[width=\textwidth]{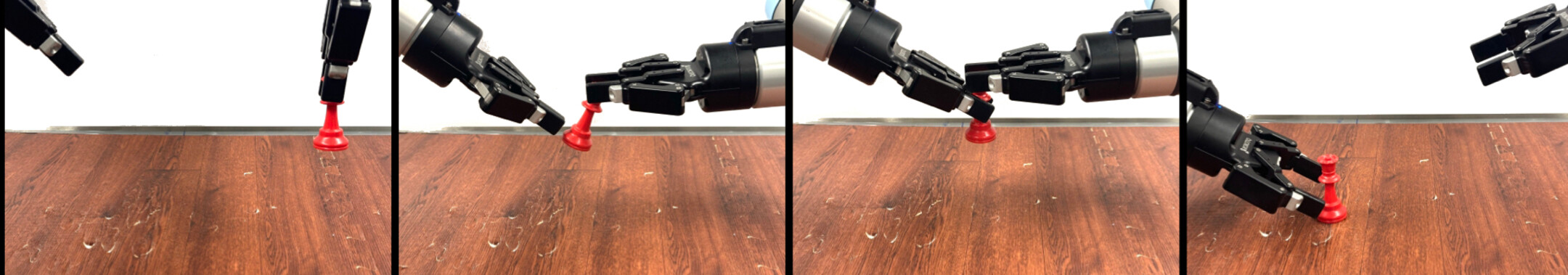}
         \caption{Handover chess piece}
         \label{fig:handover}
     \end{subfigure}
        \caption{Sequence of bimanual operations on all the 5 tasks described in the Table.~\ref{tab:task_table} are shown here.}
        \label{fig:five_tasks}
\end{figure}

Each of the teleoperation methods (3 methods for the in-person scenario and 4 methods for the remote scenario) are tested on 5 multistep bimanual manipulation tasks. These tasks are designed to include several key properties such as \textit{positional precision, rotational precision, large movements, bimanual collaboration}. The tasks are listed in the Table \ref{tab:task_table} and are illustrated in Figure \ref{fig:five_tasks}. Here \textit{positional and rotational precision} captures the need for a task to have fine movements with precision, \textit{large movements} captures the task's need for coarse movements throughout the workspace, \textit{bimanual collaboration} captures the need for with both arms to simultaneously manipulate an object during a portion of the task. In this section, we describe each of these 5 tasks and the results from our study. 

We measure the completion time of a task, along with the number of e-stops. Emergency stops are often triggered by the UR5e robots with an unsafe force, usually above 25N. While cautious teleoperation can reduce E-stops, it significantly increases completion time. In this study, the operator prioritized a competitive completion time while keeping e-stops reasonably low.

\begin{table}
\centering
\setlength\tabcolsep{0.5pt} 
\caption{\footnotesize{Key operational properties for each of the 5 bimanual manipulation tasks used in our study to evaluate the effectiveness of the SpaceMouse, VR controller, SPARK, SPARK-Remote, and its variants. }}
    \label{tab:task_table}
\begin{tabular}{ |c|c|c|c|c| }
     \hline 
     \multirow{3}{*}{Tasks} & \multicolumn{4}{c|}{Task Properties for Teleoperation} \\ \cline{2-5}
          & Positional & Large & Rotational & Bimanual \\
          & Precision & Movements & Precision & Collaboration\\
     \hline 
     Pick-and-place chess pieces      & \checkmark & \checkmark & - &- \\
     Unzip-and-open bag      & \checkmark  & - & \checkmark & \checkmark \\
     Pick-and-put markers in bag    & \checkmark & \checkmark &  \checkmark & - \\
     Untie a knot       & - & - &  \checkmark & \checkmark\\
     Handover chess piece       & \checkmark & - &  - & \checkmark\\
     \hline
\end{tabular}
\end{table}

\begin{figure*}[!t]%
    \centering
    \includegraphics[width=0.9\textwidth]{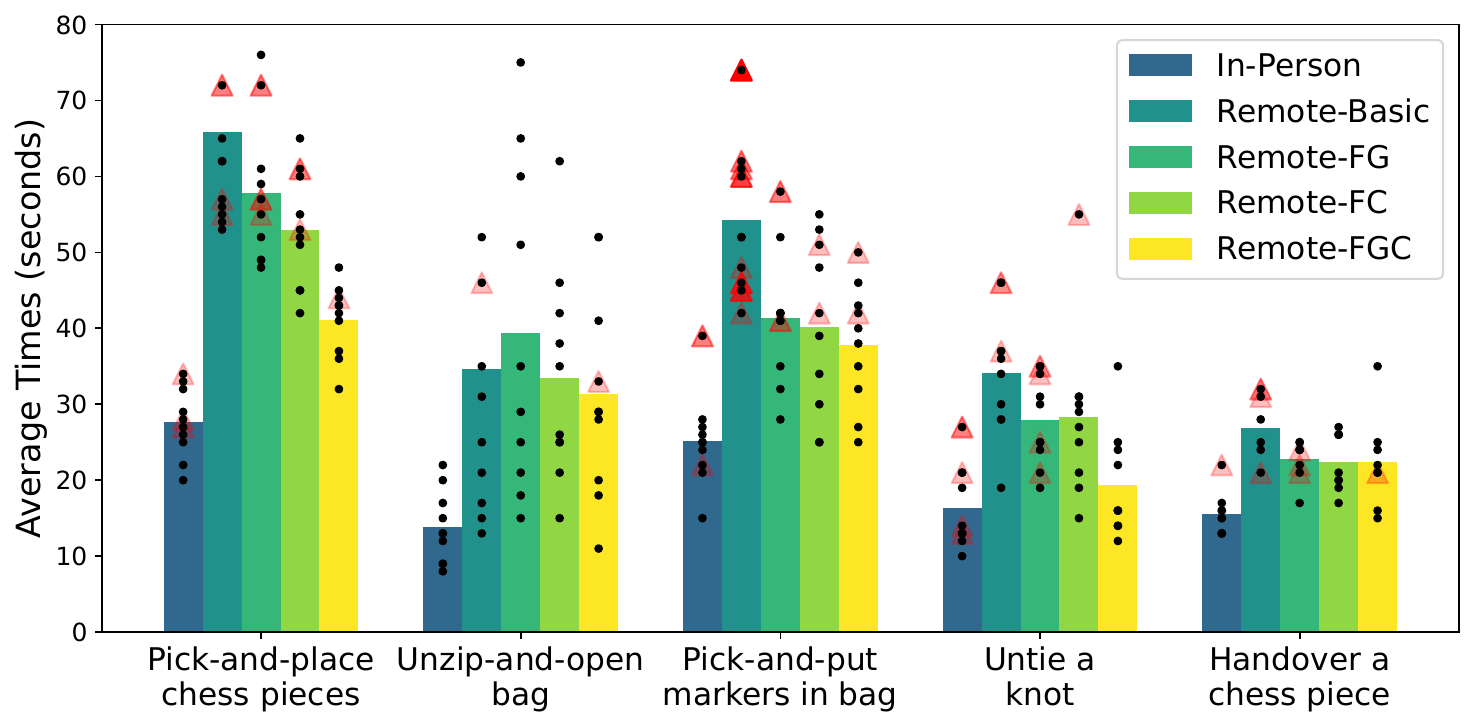} 
    \caption{Average task completion times across five tasks. Each task is evaluated 10 times, where the completion time and number of e-stops is collected. E-stops are represented as red triangles. In-Person: The operator can see the manipulators, Remote-Basic: The operator is in a remote location with a 2D video stream, Remote-FG: The operator has a screen and a force glove, Remote-FC: The operator has a screen and the manipulator has force compliance, Remote-FGC: The operator has a screen the force glove and the manipulator has force compliance. Note the use of Remote- instead of SPARK-Remote- for brevity.}%
    \label{fig:results_avg_times}%
\end{figure*}

\begin{figure}[t]%
    \centering
    \includegraphics[width=1.0\columnwidth]{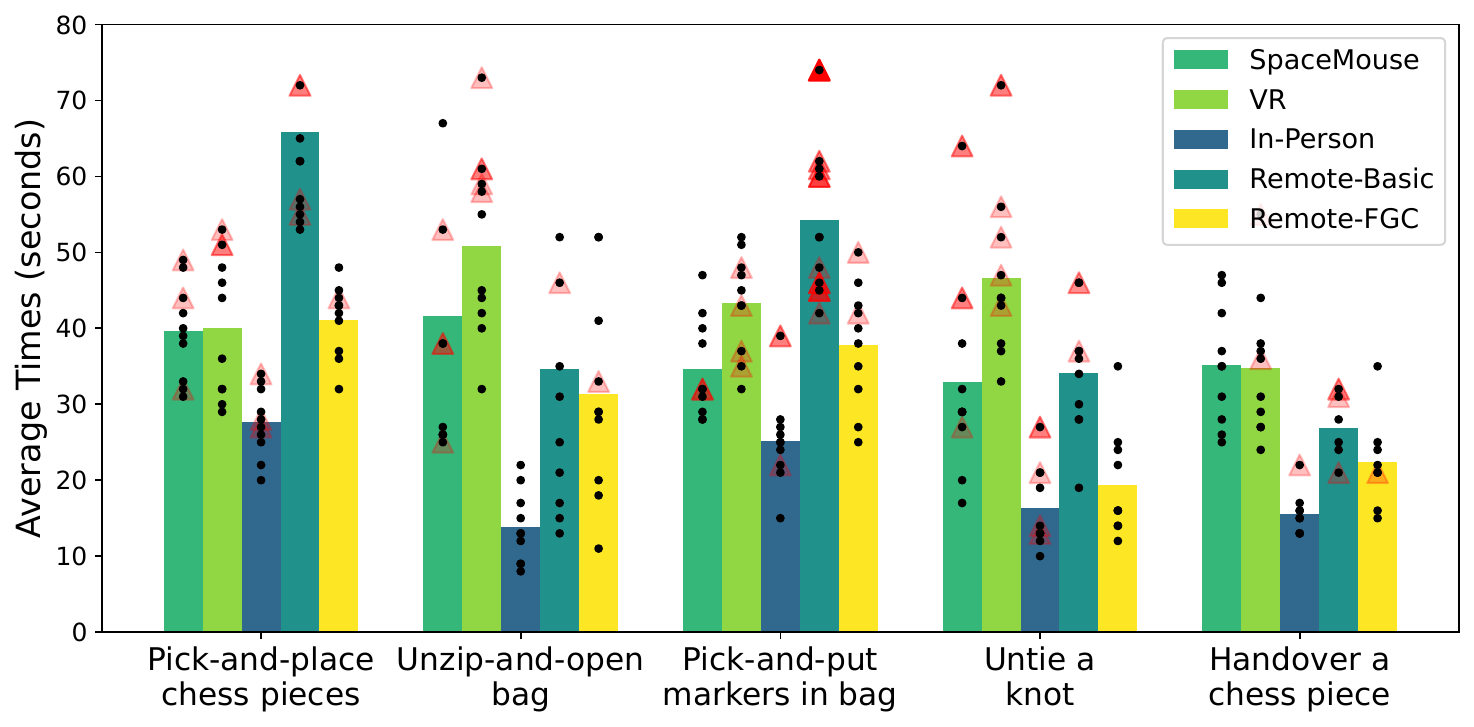} 
    \caption{An in-person evaluation compares SPARK with commercial teleoperation solutions, including VR controllers and the SpaceMouse. Results show that SPARK in-person outperforms both alternatives (aligning with findings from GELLO~\cite{gello}). Transitioning to Remote-Basic increases task completion time, while integrating the force glove (FG) and force controller (FC) in Remote-FGC reduces completion time and significantly decreases e-stops, bringing performance closer to in-person SPARK.}%
    \label{fig:results_in_person_times}%
\end{figure}

\begin{center}
    \textbf{Pick-and-place chess pieces}
\end{center}

The chess task (Figure~\ref{fig:chess_task}) begins with 6 pieces in the center (3 red and 3 green) and 2 of each color on the sides. 
This task simulates a two-dimensional pick-and-place scenario that requires no rotation, emphasizing the attributes of \textit{positional precision} and \textit{large movements} in a task. 

For the in-person teleoperation scenario, Figure~\ref{fig:results_in_person_times} shows that the SpaceMouse and VR perform decently well with positionally precise and large $(x, y, z)$ movements, making them strong choices for 2D pick-and-place tasks. SPARK-Remote-FGC performs nearly as well as the SpaceMouse and VR, indicating that network performance can be similarly intuitive to using one of those devices in-person. 

\textbf{Completion time:} Transitioning from in-person SPARK to remote operation with SPARK-Remote-Basic, we observed that task completion time more than doubled on average, increasing from 27.6s to 65.8s (see Figure~\ref{fig:results_avg_times}). This increase is primarily due to the loss of depth perception from relying only on a single camera view. Introducing force feedback with SPARK-Remote-FG reduced the completion time by 8.0s as the operator is able to tell when the piece is making a contact with the table, increasing their confidence in contact perception. The force compensation with SPARK-Remote-FC also helps reduce the completion time by 12.9s from Remote-Basic, as the operator does not need to focus as hard on avoiding collisions, allowing a controlled contact with the table. Together, the combined Remote-FGC setup achieved the fastest completion time, reducing completion times by 24.7s (41.1s total). 

\textbf{Number of E-stops:} With this task, e-stops were most common when placing chess pieces, as the piece will transfer the full force of the robot into the table if the operator is not careful. This is a common occurrence for this task, as there are six pieces to move, and the goal is to move fast. We find that the methods with force compensation (SPARK-Remote-FC, SPARK-Remote-FGC) are effective as they have less e-stops and faster compared to other remote methods.

\begin{center}
    \textbf{Unzip-and-open the bag}
\end{center}

This unzipping task (shown in Figure.~\ref{fig:unzip}) involves fully unzipping a pouch placed at the center of the workspace. This task requires \textit{positional and rotational precision} to grab the 16mm diameter pull-tab, demonstrating each method’s capability for fine manipulation. To complete this task, one arm must hold the bag, while the other unzips it (\textit{bimanual collaboration}).

For the in-person teleoperation scenario, SPARK outperforms SpaceMouse and VR, by a significant margin. The difficulty is in perceiving and grasping the pull-tab and zipper, where both the SpaceMouse and VR struggle with precise movements and rotations (see Figure~\ref{fig:results_in_person_times}). 

\textbf{Completion time:}
Moving from the in-person SPARK to SPARK-Remote versions significantly increases the completion time, and the remote versions show a lot of variance across the trials (see Figure~\ref{fig:results_avg_times}). This is due to the challenge in perceiving and grasping the pull-tab and zipper, which is amplified by the lack of depth perception. Because of this wide range of results, it is difficult to find trends within the remote variants.

\textbf{Number of E-stops:} The e-stops are not common in this task, as the bag is a deformable object with no rigid structures during manipulation. This lack of rigid contacts also may explain why there are no meaningful improvements with the glove and the compliance (with Remote-FG, FC and FGC variants).

\begin{center}
    \textbf{Pick-and-put markers in a bag}
\end{center}

This pick-and-put task begins with an open marker pouch positioned on its side and four markers on the table (shown in Figure.~\ref{fig:markers}). The task is complete when the markers are in the bag. This task challenges the operator’s ability to efficiently position and rotate the gripper to pick up the markers and put them into the narrow opening of the pouch, thus requiring \textit{positional, rotational precision} and \textit{large movements}. It is important to note that picking up the markers is significantly harder than picking up chess pieces, as the markers are only 10 mm tall, compared to the approximately 60 mm chess pieces.

SPARK provides an intuitive interface for rotating the bag and precisely grasping the markers, enabling the operator to create human-like trajectories. In contrast, with the VR and SpaceMouse setups, more of the time was spent reorienting the grippers, resulting in less effective demonstrations (see Figure~\ref{fig:results_in_person_times}). 

\textbf{Completion time:} 
Transitioning from in-person SPARK to remote operation with SPARK-Remote-Basic, we observe a doubling of task completion time, increasing from 25.2s to 54.2s (see Figure~\ref{fig:results_avg_times}). This increase was primarily due to loss of depth perception and arm occlusion in the third-person camera view, leading to frequent table contacts and e-stops. SPARK-Remote-FG, incorporating force feedback, reduced completion time by 12.9s, enabling the operator to react to unintended table contact. SPARK-Remote-FC, with force compensation, reducing the time by 14s from SPARK-Remote-Basic, allowing the user to intentionally contact the table to find the depth of the grasp. The SPARK-Remote-FGC setup achieved the fastest completion time, improving 16.4s over Remote-Basic, as combining force feedback and force compensation provided multiple strategies to complete the task efficiently.

\textbf{Number of E-stops:}
The high number of e-stops in SPARK-Remote-Basic resulted from frequent table contact, as mentioned earlier. This required the operator to proceed cautiously, as even minor errors could trigger an e-stop during the pick operation. Both the force gloves and force controller prove highly beneficial for mitigating these challenges in the remote setting.

\begin{center}
    \textbf{Untie a Knot}
\end{center}

This untieing task starts with a figure 8 knot in a rope one foot from the end, and is completed when the knot is fully untied (see Figure.~\ref{fig:untie}). Both arms must work together (\textit{bimanual collaboration}) to loosen the knot and remove the tail end (\textit{rotational precision}). The difficulty of this task is due to the proximity of both end effectors, as one usually holds a section of the knot while the other loosens the free end. 

SPARK excelled in this task by replicating natural hand movements, enabling efficient bimanual manipulation to untie the knot. In contrast, the SpaceMouse and VR setups took longer to achieve the necessary rotations (see Figure~\ref{fig:results_in_person_times}).

\textbf{Completion time:} 
Transitioning from in-person SPARK to remote operation with SPARK-Remote-Basic, task completion time doubled from 16.3s to 34.1s (see Figure~\ref{fig:results_avg_times}). While depth perception plays a role, video compression can also obscure the knot’s geometry due to the rope's uniform texture. Force feedback in SPARK-Remote-FG made it 6.2s faster than Remote-Basic, and SPARK-Remote-FC completed 5.8s faster than Remote-Basic. The Remote-FGC setup achieved the fastest completion, reducing Remote-Basic's time by 14.7s, as force feedback and force compliance enabled the operator to leverage contact for faster task execution (only 3.1s slower than in-person).

\textbf{Number of E-stops:}
Although this task involves a deformable rope, the proximity of the end effectors increases the risk of gripper-gripper contact. This risk can be minimized with slower, cautious strategies or embraced by taking more precise but potentially riskier trajectories. SPARK-Remote-Basic requires a conservative approach, as e-stops are highly likely otherwise. The force glove (SPARK-Remote-FG) reduce the consequences of contact, though mistakes can still trigger e-stops. The compliant method (SPARK-Remote-FC) enables a similar approach, but significantly reduces e-stops due to its tolerance for errors. SPARK-Remote-FGC offers the greatest operational confidence, leveraging force feedback to enhance force compliance while eliminating e-stops.

\begin{center}
    \textbf{Handover chess piece}
\end{center}

This handover task has the operator pick up a chess piece with one arm, complete an in-gripper rotation of the piece using the other arm, hand it to the other arm, and place it on the table. (see Figure.~\ref{fig:handover}).
This task emphasizes contact-rich interactions, hand-to-hand transfer, and precise placement, showcasing: \textit{positional precision} and \textit{bimanual collaboration}. 

We find that the SpaceMouse and VR perform adequately, but all variants of SPARK provide a faster task completion as the bimanual motions are more natural (see Figure~\ref{fig:results_in_person_times}).

\textbf{Completion time:} 
The task could be completed in 15.6s in person, and SPARK-Remote-Basic took 26.8s on average, 11.2s behind the in person setup (see Figure~\ref{fig:results_avg_times}). The other remote variations took around 22 seconds, only improving by about 4 seconds. We find that the task does not provide a significant difference in completion times within the remote setting. 

\textbf{Number of E-stops:}
This task contains several contact-rich subtasks: in-gripper manipulation, handover, and place operations, with the handover presenting the most challenge. The handover requires the grippers to be nearly perfectly aligned to avoid an e-stop. Force compliance allowed both Remote-FC and Remote-FGC to complete the task with only 1 e-stop between them.  Remote-Basic and Remote-FG do not have force compliance, so mistakes generally result in e-stops, with 6 between both methods.


%% file: sections/3_Results/Discussion.tex
\section{Discussion}

The results of this study show that SPARK-Remote-FGC, a cost-effective and kinematically scaled teleoperation system, significantly enhances the effectiveness of bimanual remote robot teleoperation. The system utilizes haptic feedback through a force glove and force control via a compensation algorithm, both of which address key challenges in remote operation. We observe that moving from in-person to remote teleoperation significantly impacts performance due to loss of depth perception, increased contact uncertainty, and a higher risk of unintended high-force interactions.

Our evaluation of SPARK-Remote and its variants across five bimanual manipulation tasks provides several insights into the role of teleoperation interfaces in task performance. Transitioning from in-person teleoperation to remote operation results in a significant performance drop due to the absence of direct visual feedback and depth cues. However, through the addition of haptic feedback (SPARK-Remote-FG) and force control (SPARK-Remote-FC), we can improve both task completion time and reducing emergency stops (e-stops). The combination of both enhancements (SPARK-Remote-FGC) yielded the best results across all tasks, demonstrating the complementary benefits of force sensing and adaptive force regulation.

While SPARK-Remote-FGC shows considerable improvements in remote teleoperation, some limitations remain. The reliance on a single fixed camera view in the remote setup constrains the operator’s perspective, potentially limiting the usefulness of force feedback in tasks requiring precise spatial awareness. Future iterations could explore multi-camera setups or depth-aware (RGBD) visualization techniques to further enhance remote operator awareness. Additionally, while the force glove provides valuable tactile feedback, it lacks the intuitiveness of high-fidelity haptic devices, which could be explored through the development and evaluation of low-cost tactile feedback devices.

Another key observation is that SPARK-Remote-FGC has inconsistent performance improvements across our tasks. For example, in tasks with rigid-object manipulation, such as pick-and-place, force feedback significantly improved precision and speed. However, for deformable-object manipulation tasks like unzipping a bag, force feedback has limited benefits due to the highly compliant nature of the object. This suggests that future designs could incorporate adaptive feedback mechanisms that adjust the force glove's feedback sensitivity depending on the task.

Beyond task performance, SPARK-Remote offers a cost-effective alternative to high-end bilateral teleoperation systems. Compared to commercially available solutions, which often exceed \$40,000 per arm, SPARK-Remote achieves high performance at a fraction of the cost (\$200 per arm). This affordability makes SPARK-Remote a viable option for broader adoption in applications such as assistive robotics, hazardous environment operations, and data collection.

%% file: sections/4_Conclusion/Conclusion.tex
\section{Conclusion}
\label{sec:conclusion_chapter}

This work introduces SPARK-Remote, a low-cost and effective teleoperation system for bimanual robots, addressing key challenges in remote robot control. Through a comparative evaluation against existing teleoperation technologies and a structured set of bimanual tasks, the system demonstrates significant performance benefits, particularly when incorporating force feedback and force control mechanisms. The findings highlight that while remote teleoperation inherently presents challenges, haptic feedback and adaptive force regulation can bridge the gap between in-person and remote operation.

The success of SPARK-Remote suggests promising directions for future teleoperation research. Expanding the system to include multi-modal feedback (e.g., audio cues, visual overlays) and enhanced force modeling could further improve its usability and performance. Additionally, its open-source nature paves the way for further refinements and community-driven improvements.

In conclusion, SPARK-Remote provides a practical, scalable, and cost-effective solution for improving bimanual robot teleoperation, particularly in scenarios where traditional high-cost systems are impractical. By combining affordability with intuitive control enhancements, this work contributes to the ongoing effort to make robot teleoperation more accessible, efficient, and user-friendly across diverse applications.

%% file: sections/4_Conclusion/Acknowledgments.tex
\section{Acknowledgments}
\label{sec:Acknowledgments_chapter}

The authors would like to thank Carl Winge for supporting the hardware development, the Minnesota Robotics Institute for providing funding through the MnRI Seed Grant, and the University of Minnesota's Research \& Innovation Office for providing funding through the Early Innovation Fund.

%% file: spark.bbl
\begin{thebibliography}{10}
\providecommand{\url}[1]{#1}
\csname url@samestyle\endcsname
\providecommand{\newblock}{\relax}
\providecommand{\bibinfo}[2]{#2}
\providecommand{\BIBentrySTDinterwordspacing}{\spaceskip=0pt\relax}
\providecommand{\BIBentryALTinterwordstretchfactor}{4}
\providecommand{\BIBentryALTinterwordspacing}{\spaceskip=\fontdimen2\font plus
\BIBentryALTinterwordstretchfactor\fontdimen3\font minus \fontdimen4\font\relax}
\providecommand{\BIBforeignlanguage}[2]{{%
\expandafter\ifx\csname l@#1\endcsname\relax
\typeout{** WARNING: IEEEtran.bst: No hyphenation pattern has been}%
\typeout{** loaded for the language `#1'. Using the pattern for}%
\typeout{** the default language instead.}%
\else
\language=\csname l@#1\endcsname
\fi
#2}}
\providecommand{\BIBdecl}{\relax}
\BIBdecl

\bibitem{diff}
C.~Chi, Z.~Xu, S.~Feng, E.~Cousineau, Y.~Du, B.~Burchfiel, R.~Tedrake, and S.~Song, ``Diffusion policy: Visuomotor policy learning via action diffusion,'' \emph{The International Journal of Robotics Research}, p. 02783649241273668, 2023.

\bibitem{shared}
H.~Liu, S.~Nasiriany, L.~Zhang, Z.~Bao, and Y.~Zhu, ``Robot learning on the job: Human-in-the-loop autonomy and learning during deployment,'' \emph{The International Journal of Robotics Research}, p. 02783649241273901, 2022.

\bibitem{chu2023bootstrapping}
X.~Chu, Y.~Tang, L.~H. Kwok, Y.~Cai, and K.~W.~S. Au, ``Bootstrapping robotic skill learning with intuitive teleoperation: Initial feasibility study,'' in \emph{International Symposium on Experimental Robotics}.\hskip 1em plus 0.5em minus 0.4em\relax Springer, 2023, pp. 42--52.

\bibitem{gello}
P.~Wu, Y.~Shentu, Z.~Yi, X.~Lin, and P.~Abbeel, ``Gello: A general, low-cost, and intuitive teleoperation framework for robot manipulators,'' in \emph{2024 IEEE/RSJ International Conference on Intelligent Robots and Systems (IROS)}.\hskip 1em plus 0.5em minus 0.4em\relax IEEE, 2024, pp. 12\,156--12\,163.

\bibitem{peract}
M.~Shridhar, L.~Manuelli, and D.~Fox, ``Perceiver-actor: A multi-task transformer for robotic manipulation,'' in \emph{Conference on Robot Learning}.\hskip 1em plus 0.5em minus 0.4em\relax PMLR, 2023, pp. 785--799.

\bibitem{seo2023deep}
M.~Seo, S.~Han, K.~Sim, S.~H. Bang, C.~Gonzalez, L.~Sentis, and Y.~Zhu, ``Deep imitation learning for humanoid loco-manipulation through human teleoperation,'' in \emph{2023 IEEE-RAS 22nd International Conference on Humanoid Robots (Humanoids)}.\hskip 1em plus 0.5em minus 0.4em\relax IEEE, 2023, pp. 1--8.

\bibitem{aloha}
T.~Z. Zhao, V.~Kumar, S.~Levine, and C.~Finn, ``Learning fine-grained bimanual manipulation with low-cost hardware,'' \emph{arXiv preprint arXiv:2304.13705}, 2023.

\bibitem{mobile-aloha}
Z.~Fu, T.~Z. Zhao, and C.~Finn, ``Mobile aloha: Learning bimanual mobile manipulation using low-cost whole-body teleoperation,'' in \emph{8th Annual Conference on Robot Learning}, 2024.

\bibitem{kim2023training}
H.~Kim, Y.~Ohmura, A.~Nagakubo, and Y.~Kuniyoshi, ``Training robots without robots: deep imitation learning for master-to-robot policy transfer,'' \emph{IEEE Robotics and Automation Letters}, vol.~8, no.~5, pp. 2906--2913, 2023.

\bibitem{yangace}
S.~Yang, M.~Liu, Y.~Qin, R.~Ding, J.~Li, X.~Cheng, R.~Yang, S.~Yi, and X.~Wang, ``Ace: A cross-platform and visual-exoskeletons system for low-cost dexterous teleoperation,'' in \emph{8th Annual Conference on Robot Learning}.

\bibitem{tele}
S.~Lichiardopol, ``A survey on teleoperation,'' 2007.

\bibitem{VR}
A.~Naceri, D.~Mazzanti, J.~Bimbo, D.~Prattichizzo, D.~G. Caldwell, L.~S. Mattos, and N.~Deshpande, ``Towards a virtual reality interface for remote robotic teleoperation,'' in \emph{2019 19th International Conference on Advanced Robotics (ICAR)}, 2019, pp. 284--289.

\bibitem{comp}
\BIBentryALTinterwordspacing
T.~Kamijo, C.~C. Beltran-Hernandez, and M.~Hamaya, ``Learning variable compliance control from a few demonstrations for bimanual robot with haptic feedback teleoperation system,'' 2024. [Online]. Available: \url{https://arxiv.org/abs/2406.14990}
\BIBentrySTDinterwordspacing

\bibitem{glove}
\BIBentryALTinterwordspacing
E.~Rueckert, R.~Lioutikov, R.~Calandra, M.~Schmidt, P.~Beckerle, and J.~Peters, ``Low-cost sensor glove with force feedback for learning from demonstrations using probabilistic trajectory representations,'' 2015. [Online]. Available: \url{https://arxiv.org/abs/1510.03253}
\BIBentrySTDinterwordspacing

\bibitem{Zhou2021ABD}
\BIBentryALTinterwordspacing
C.~Zhou, L.~Zhao, H.~Wang, L.~Chen, and Y.~Zheng, ``A bilateral dual-arm teleoperation robot system with a unified control architecture,'' \emph{2021 30th IEEE International Conference on Robot \& Human Interactive Communication (RO-MAN)}, pp. 495--502, 2021. [Online]. Available: \url{https://api.semanticscholar.org/CorpusID:237296363}
\BIBentrySTDinterwordspacing

\bibitem{9568842}
C.~Lenz and S.~Behnke, ``Bimanual telemanipulation with force and haptic feedback and predictive limit avoidance,'' in \emph{2021 European Conference on Mobile Robots (ECMR)}, 2021, pp. 1--7.

\bibitem{xu2025immersive}
H.~Xu, M.~Chen, G.~Li, L.~Wei, S.~Peng, H.~Xu, and Q.~Li, ``An immersive virtual reality bimanual telerobotic system with haptic feedback,'' \emph{arXiv preprint arXiv:2501.00822}, 2025.

\end{thebibliography}
